\documentclass[fleqn,10pt]{wlscirep}
\usepackage[utf8]{inputenc}
\usepackage[T1]{fontenc}
\usepackage{tabularx}
\usepackage{graphicx}
\usepackage{orcidlink}

\title{PetroGAN: A novel GAN-based approach to generate realistic, label-free petrographic datasets}

\author[1,+]{Iván Ferreira\,\orcidlink{0000-0001-9369-2189}}
\author[1]{Luis Ochoa\,\orcidlink{0000-0002-3607-7339}}
\author[2,*,+]{Ardiansyah Koeshidayatullah\,\orcidlink{0000-0003-0940-678X}}
\affil[1]{Departamento de Geociencias, Universidad Nacional de Colombia, Bogotá, Colombia}
\affil[2]{Department of Geosciences, College of Petroleum and Minerals , King Fahd University of Petroleum and Minerals, Dhahran, Saudi Arabia}

\affil[*]{a.koeshidayatullah@kfupm.edu.sa}

\affil[+]{these authors contributed equally to this work}

\begin{abstract}
Deep learning architectures have enriched data analytics in the geosciences, complementing traditional approaches to geological problems. Although deep learning applications in geosciences show encouraging signs, the actual potential remains untapped. This is primarily because geological datasets, particularly petrography, are limited, time-consuming, and expensive to obtain, requiring in-depth knowledge to provide a high-quality labeled dataset. We approached these issues by developing a novel deep learning framework based on generative adversarial networks (GANs) to create the first realistic synthetic petrographic dataset. The StyleGAN2 architecture is selected to allow robust replication of statistical and esthetical characteristics, and improving the internal variance of petrographic data. The training dataset consists of 10070 images of rock thin sections both in plane- and cross-polarized light. The algorithm trained for 264 GPU hours and reached a state-of-the-art Fréchet Inception Distance (FID) score of 12.49 for petrographic images. We further observed the FID values vary with lithology type and image resolution. Our survey established that subject matter experts found the generated images were indistinguishable from real images. This study highlights that GANs are a powerful method for generating realistic synthetic data, experimenting with the latent space, and as a future tool for self-labelling, reducing the effort of creating geological datasets.
\end{abstract}

\keywords{petrography, StyleGAN, generative adversarial network, deep learning, thin section}

\begin{document}

\flushbottom
\maketitle

\section*{Introduction}

Advances in artificial intelligence and machine learning have been used effectively in the last decades to analyze and generate meaningful insights from geological data, using a vast array of available algorithms \cite{Cate2017, Mosser2017, Dramsch2020, Koeshidayatullah2020}. Recently, with the advent of generative models like Generative Adversarial Networks (GANs) \cite{Goodfellow2014}, Variational Auto-Encoders (VAEs) \cite{Kingma2013} ,transformer GANs \cite{Jiang2021} and Diffusion models \cite{Dhariwal2021} there is an enormous potential to further explore the application of this deep learning technique in the geosciences. 

GANs have been shown to generate realistic and diverse images in an unsupervised manner and are already adopted in several fields, including super-resolution, Image-to-Image translation, Text-to-Image translation, Style-Mixing, and generation of realistic images   and are positioned amongst the most useful generative models. In general, GANs aim is to capture the data distribution via a minimax two-player game that aims to produce synthetic samples based on the original dataset, mimicking its statistical and esthetical characteristics \cite{Goodfellow2014} even going as far as deceiving human observers in the ability to discriminate real images from generated ones \cite{Curto2017, Lago2022, Nightingale2022}.

Geosciences in general have been adopting deep learning-based analytics in its workflows, such as image processing tasks. However, the lack of labeled, varied, and sufficiently large datasets \cite{Izadi2020} have resulted in images being overtrained and overfit to certain geological contexts \cite{Pires2020} or with deep learning algorithms such as Convolutional Neural Networks (CNNs) with not enough data to yield satisfactory result \cite{Maitre2019}.The use of transfer learning has been suggested as an alternative \cite{Koeshidayatullah2020, Pires2021, Wu2020}, with the risk of overtraining on a single geological context using this kind of approach. Furthermore, the high accuracy obtained from the transfer learning methods creates another dimension of uncertainty whereby a model trained to recognize animals or other daily objects can be applied to classify geological images, such as seismic and petrographic images that have entirely different high- and low-level features. Therefore, there is still a gap that generative models could help address by bringing deep learning applications closer to geological tasks related to image processing. 

Although historically clustering techniques and edge detection algorithms were the main algorithms for digital image processing \cite{Koh2021}, recent studies have started to explore and utilize deep CNNs to address visual recognition and image processing in geosciences, including petrographic images \cite{Koeshidayatullah2020, Koh2021}. To date, the implementation of GANs in the petrographic image dataset has been limited. One of the major limitations to applying GANs in petrographic images and others geoscience datasets is that most GAN architectures require massive amounts of data, as illustrated in the first implementation of a GAN with the MNIST and CIFAR datasets (70,000 images each). \cite{Goodfellow2014}. Furthermore, the use of image datasets in the 1000s range has been shown to lead to the overfitting of the generator in GANs \cite{Feng2022}. 
This work addresses the implementation of GANs in a petrographic context by (i) experimenting with different dataset sizes, (ii) using various image resolutions, and (iii) applying truncation values with the aim of creating a framework to generate petrographic datasets. Furthermore, our work uses the Adaptive Discriminator Augmentation (ADA) implementation of StyleGAN2 \cite{Karras2018, Karras2019, Karras2020} to realistically generate synthetic petrographic images due to the high quality of generated images by this model, backward compatibility, stability, and the viability of its use in nonfacial generation datasets. In addition, the implementation of the state-of-the-art Fréchet Inception Distance (FID)\cite{Heusel2017} scores provide a better metric to evaluate the generated images. The dataset size limitation in petrographic thin sections is a problem addressed in this paper, as images collected and sliced were of a sufficient volume after preprocessing to be able to produce meaningful results using this kind of generative model.  The ultimate objectives of this work are to (i) explore the best image resolution and dataset size to generate realistic thin sections, (ii) develop a novel deep learning framework to generate petrographic synthetic datasets, and (iii) discuss the properties of a petrographic GAN model using latent space, transfer learning, interpolation, truncation, and feature extraction\cite{Brock2018, Karras2020, Shen2020, Radford2016}, and (iv) evaluate the synthetic datasets through a simple survey asking opinions from subject matter experts. We further aim to highlight the implementation of GAN algorithms and other generative models as a way forward for exploring self-labeling and image generation tasks, and it could serve to support the implementation of deep learning algorithms for image analysis applied to petrography in future workflows.

\section*{Related Work}

GANs have been recently adopted in geosciences with the motivations to explore and apply generative models as a way of generating and manipulating a latent space associated with the geological data of interest, i.e., the high-dimensional space where a representation of the data is encoded \cite{White2016, Mosser2017, Song2021}. This space is used to upscale the dimensionality and upsample the quality. Previous works have demonstrated the far-reaching impact and application of GANs in geosciences, from reservoir simulation to history matching (see Table \ref{tab:table1}). \cite{Mosser2017,Coiffier2020, Niu2020, Jo2021, Song2021}

\begin{table}[ht]
\centering
\begin{tabular}{|l|l|l|}
\hline
Reference & GAN Algorithm used & Problem Addresed \\
\hline
\cite{Mosser2017} & Volumetric DCGAN & Reconstruction of porous media from images of sedimentary rocks. \\
\hline
\cite{Coiffier2020} & Dimension Augmenter GAN & Generate 3D stochastic fields from 2D images for hydrology. \\
\hline
\cite{Nanjo2020} & Not Specified & Reconstruction and classification of carbonate thin sections. \\
\hline
\cite{Niu2020} & Cycle-in-Cycle GAN & Improve the resolution of 3D micro-CT images. \\
\hline
\cite{Song2021} & Progressive growth GAN & Generation of 2D geological facies models. \\
\hline
\cite{Jo2021} & GAN and Ensemble Kalman filter & Assisted history matching for a Deepwater lobe system. \\
\hline
\end{tabular}
\caption{\label{tab:table1}Use of GANs in geological datasets.}
\end{table}

Furthermore, GANs have been proposed as a tool to create synthetic carbonate components \cite{Koeshidayatullah2020} and for obtaining super-resolution micro-computed tomography (Micro-CT) images \cite{Niu2020} for digital rock physics workflows. Furthermore, GANs have also been used successfully to assist in the reconstruction and classification of carbonate thin sections \cite{Nanjo2020}, positioning GANs as a possible tool to enhance carbonate lithology interpretation workflows in combination with core images and Fullbore Formation MicroImager (FMI) images.  

\section*{Methods}

In this study, the datasets consisted of RGB thin-section images in both Cross-polarized (XPL) and plane-polarized (PPL). Information from XPL and PPL images are crucial to determine the type of minerals and lithological variations in thin sections. The datasets were prepared by using (i) the provided dataset tool generation from the StyleGAN repository and (ii) image slicing as a method of data augmentation. The StyleGAN architecture was selected based on its state-of-the-art scores achieved and the ability to experiment with the generated latent space (Table \ref{tab:table2}) \cite{Karras2019,Karras2020}.

\begin{table}[ht]
\centering
\begin{tabular}{|l|l|l|l|l|}
\hline
GAN Implementation& 2k Training set & 5k Training set & 10k Training set & 140k Training set\\
\hline
BigGAN \cite{Brock2018} & 60.47 & 32.34 & 15.85 & 11.08\\
\hline
StyleGAN2 \cite{Karras2019} & 66.77 & 39.42 & 8.80 & 3.81\\
\hline
+ADA \cite{Huang2017,Karras2020}& 15.76 & 10.78 & 5.40 & 3.79\\
\hline
\end{tabular}
\caption{\label{tab:table2}Comparison of FID Scores on images generated at 256x256 for different dataset sizes from the Flickr Faces HQ Dataset (FFHQ) comparing StyleGAN2 with BigGAN, and the StyleGAN2+ADA comparing Fréchet Inception Distance (FID) Scores. Adapted from the StyleGAN2 + ADA original paper \cite{Karras2020}.}
\end{table}

\subsection*{StyleGAN and StyleGAN2 architecture}

\begin{figure}[ht]
\centering
\includegraphics[width=\linewidth]{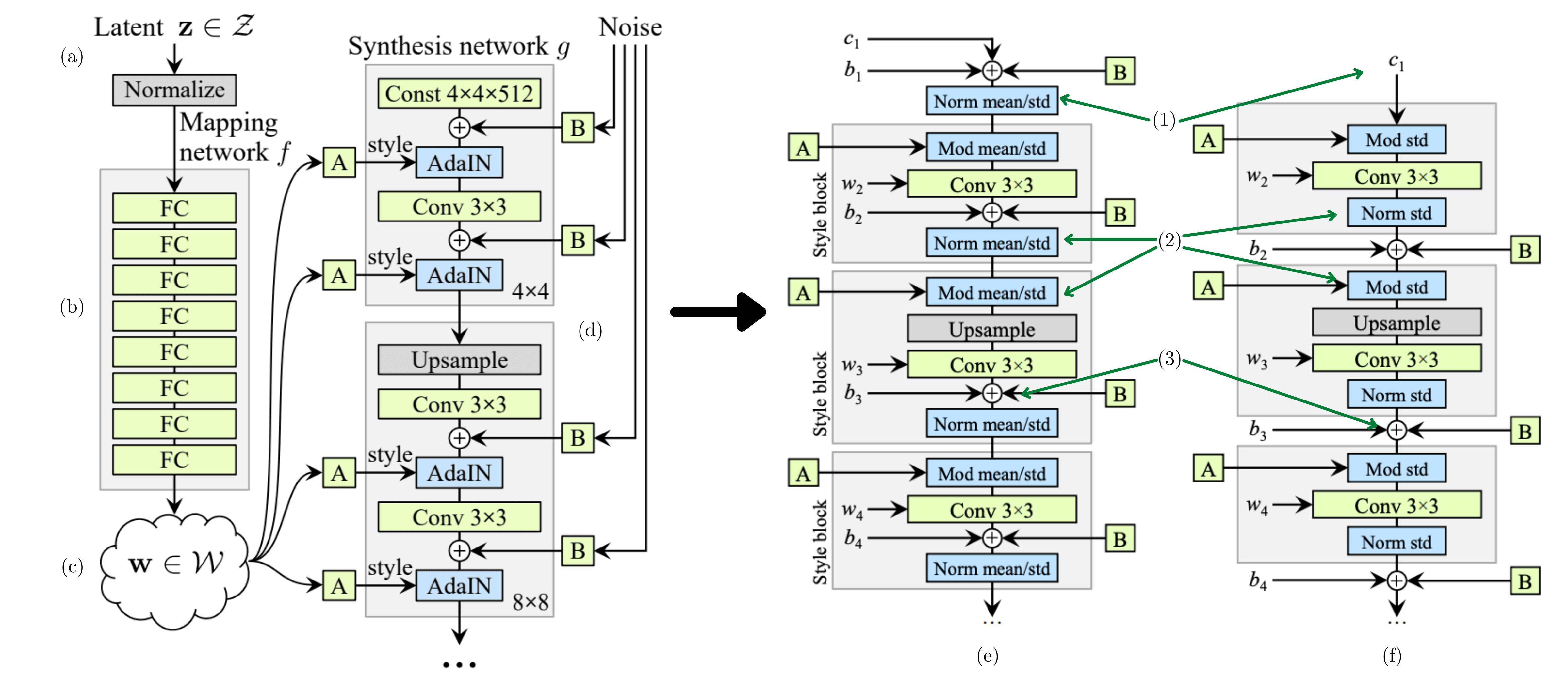}
\caption{Original StyleGAN architecture, (a) The latent vector z introduced, (b) eight fully connected layers used to obtain (c) latent code w containing the features and (d) a series of AdaIN, normalization and convolutions using progressive growth to generate high resolution images (Karras et al., 2018). Major modifications were made to (e) StyleGAN in (f) StyleGAN2. The model is simplified by removing the initial processing of the constant (1), the removal of the mean in the process of normalizing the features (2) and the transfer of the noise module (+) outside of the style block (3). Modified from the StyleGAN and StyleGAN2 papers \cite{Karras2018,Karras2019}}
\label{fig:fig1}
\end{figure}

The architecture used in this work is a style-based generator, Figure \ref{fig:fig1}. The network consists of (i) a mapping network from the latent vector z, i.e., the latent vector representation of an image in latent space Z; (ii) a mapping of this vector using eight fully connected layers into W Space, this latent code in W space, being the space of all style vectors w; is then (iii) used in conjunction with Adaptive Instance Normalization (AdaIN) layers \cite{Huang2017} to control the features in the generator. This is managed by (using a progressive growth of the image, reducing the complexity of generating high resolution images by taking a step-by-step approach \cite{Karras2018}. However, this has been linked to the generation of artifacts in the generated image, one of the main reasons behind the re-engineering of the model adopted in StyleGAN2 \cite{Karras2019, Karras2020}

For this reason, we use the second iteration of the StyleGAN line of models, Figure \ref{fig:fig1}f \cite{Karras2019,Karras2020} which further developed the original StyleGAN \cite{Karras2018}.This architecture is constantly developed and improved and has backwards compatibility with the preceding StyleGAN architectures with the same dataset preparation tools, accepted image resolutions, and workflows used. The latest iteration of this model is the StyleGAN3 \cite{Karras2021}, we did not choose this architecture because an acceptable FID Score was not achieved and the model diverged with the same dataset size. 
The generation of unintended artifacts in StyleGAN, mostly due to the progressive growth technique used, were addressed by creating the StyleGAN2 model \cite{Karras2019,Karras2020}. This was achieved by simplifying and eliminating steps in the architecture, Figure 1\ref{fig:fig1}f, and instead of using a progressive growth to increasingly generate high resolution images, the implementation of skip connections in StyleGAN2 was used \cite{Karras2019}. This method allows skipping some of the layers in the model and feeding this output to the next layers as implemented in the Residual Networks (ResNET) architectures \cite{He2016}.

\subsection*{Dataset sources}

\begin{figure}[ht]
\centering
\includegraphics[width=\linewidth]{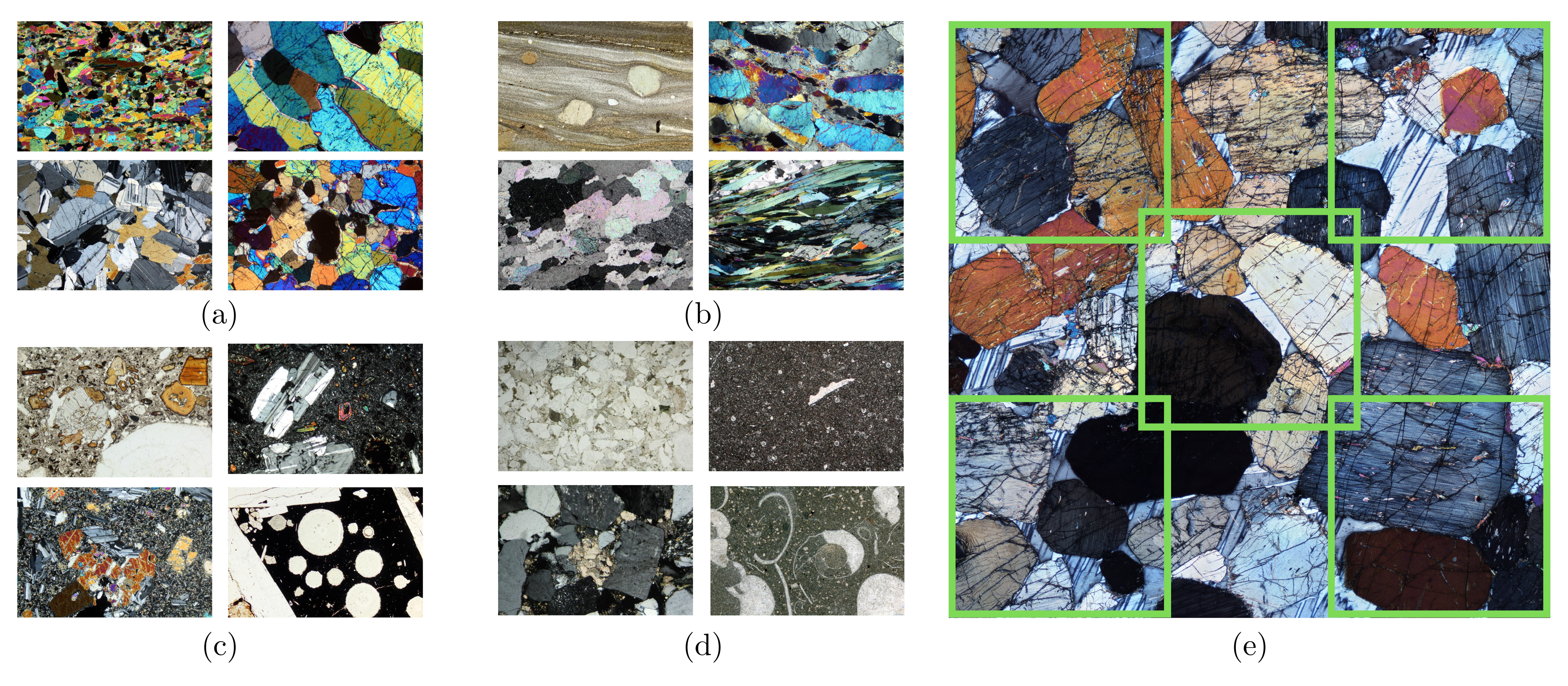}
\caption{Four lithology classes extracted from the Strekeisen dataset (a) plutonic, (b) metamorphic, (c) volcanic and (d) sedimentary rocks in thin section (e) example of image slicing applied to the VPM dataset. Image slicing was done to MacQuarie university images, splitting the original into five representative subsections.}
\label{fig:fig2}
\end{figure}

Petrographic images were collected from publicly available sources, as shown in Table 3. The dataset consists of high-resolution images 1701x1686 pixels collected from the Virtual Petrographic Microscope project (VPM) in PPL and XPL with different angles of rotation for each image \cite{Tetley2014}, Figure \ref{fig:fig2}e. This dataset is complemented by 800x533 pixel petrographic images taken from the Strekeisen project, Figure \ref{fig:fig2}a-d \cite{Mommio2007} Images from all datasets were divided into four main rock types: (1) plutonic, (2) volcanic, (3) metamorphic, and (4) sedimentary classes. Magnifications were also considered to obtain several magnifications and representations of various minerals which ranged from 10x and 20x from the Strekeisen project and the Atlas of sedimentary rocks book \cite{Adams1984}; 4x from J.D. Derochette project \cite{Derochette2008}, and full thin section photomicrographs for the VPM.

\subsection*{Image slicing and final dataset}

Data processing was performed through standard image manipulation made available as part of the StyleGAN2 implementation, Numpy \cite{Harris2020}, OpenCV, Pillow, and PyTorch \cite{Paszke2019}. As per requirement of the StyleGAN2 architecture, images needed to be in a square format with dimensions in powers of two (i.e., 32x32px, 256x256px, 512x512px, etc.). To achieve a sizeable dataset for the GAN to train on and satisfy the StyleGAN2 dataset requirement \cite{Karras2018, Karras2019, Feng2022} and using the highest possible resolution while preserving the important features of the petrographic dataset, the original images were cropped or sliced to a format of 512x512px. 

The final dataset consisted of 10070 petrographic images belonging to four classes, this combined set of images is used to train the GAN for generating 512x512px images; one of the main objectives of the generated dataset was to achieve not only a greater than 10k number of images but to have a class-balance between lithologies as shown in Table \ref{tab:table3}.

\begin{table}[htbp]
        \centering
        \small
        \setlength\tabcolsep{2pt}
\begin{tabular}{|c|c|c|c|c|}
\hline
Source & Number of images (Quantity per dataset) & Average resolution & Class & Images sliced  (512x512)\\
\hline
SP, VPM & (2549,20) & 800x533, 1701x1686 & Igneous plutonic & 2645\\
\hline
SP, VPM, JD & (1756,60,45) & 800x533,  1701x1686,  2600x1700 & Igneous volcanic & 2281\\
\hline
SP, VPM, Adams & (695,150,217) & 800x533,1701x1686, 1120x820 & Sedimentary & 2530\\
\hline
SP, VPM & (2086,132) & 800x533,1701x1686 & Metamorphic & 2614\\
\hline
\end{tabular}
\caption{\label{tab:table3}Data procedence for thin sections used for training, SP: Strekeisen Project, VPM: Virtual Microscope Project, Adams: Atlas of sedimentary rocks [Adams et al., 1984], JD: J. M. Derochette}
\end{table}

\subsection*{Metrics used to evaluate GAN performance}

\begin{figure}[ht]
\centering
\includegraphics[width=\linewidth]{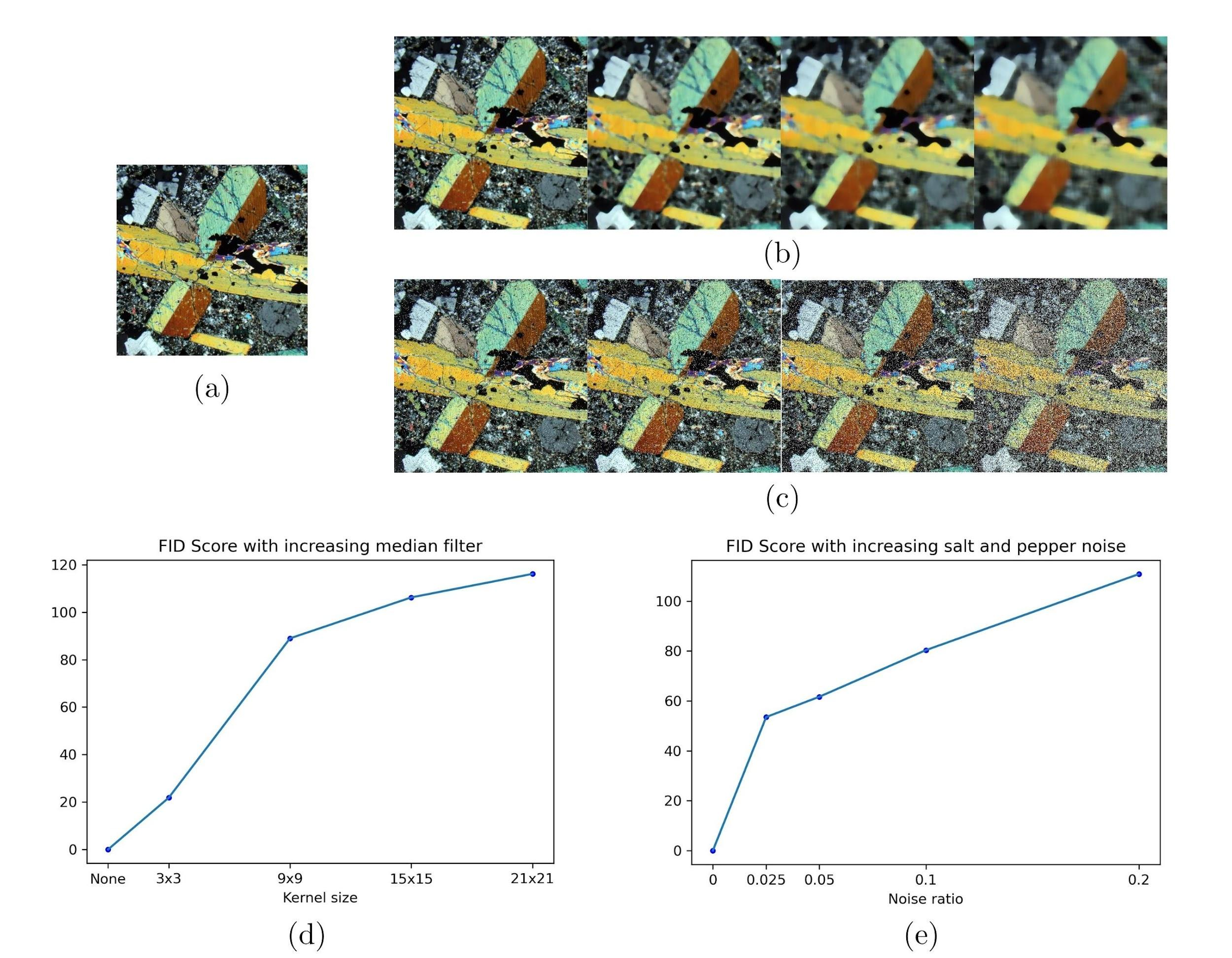}
\caption{Comparison of the effect of different disturbances on the FID Score for a nepheline foidite from the Strekeisen dataset (a) Original image (b) increasing the kernel size median filter applied to the original image (c) Adding salt and pepper noise to the image (d) Increase of the FID Score for the corresponding kernel size in the median filter (e) Increasing the noise to signal ratio effect on the FID score}
\label{fig:fig3}
\end{figure}

There are several metrics that are useful for evaluating how a GAN is performing, such as FID score, Inception score, and evaluation with domain experts. The most used and state-of-of-the-art metric is the Fréchet Inception Distance score 9,22,23,25, which is a way of capturing the similarity of generated images to real ones better than the other metrics like the Inception Score 24. In addition, this metric evaluates the statistical distribution of the generated images and how close it is to the statistical distribution of the real images, using the last layer of the InceptionV3 model \cite{Szegedy2015}to capture features of the generated and real images, summarizing the activation as a multivariate Gaussian distribution and calculating its means and covariance \cite{Heusel2017}. The distance between the distributions, real and fake, is computed using the similarity via the Fréchet distance \cite{Heusel2017}. Figure \ref{fig:fig3} illustrates the behavior of the FID score reacting to progressive image contamination in the context of petrographic images, the lower the FID score the closer two image distribution are, i.e., the closer a generated image dataset is to real images.

\subsection*{Training procedures}

\subsubsection*{Test procedure with 32x32px images}

As a Minimum Viable Product (MVP), a training was conducted using only igneous images and consisting of 15294 images of 32x32 pixels in size. Images were taken exclusively from the igneous rocks available from the VPM and SP. The objective of this test was to ensure that a convergence in the model was viable, as training time for GAN models usually needs both extensive training and high-end computing capabilities entailing one or several Graphical Processing Units (GPU). The MVP trained for four days and 13 hours, using a Quadro M4000 with 8 Gb of video RAM, 30 Gb of RAM and an eight-core CPU; the model converged and achieved an FID score of 7.49.

\subsubsection*{Training procedure for the 512x512px, 256x256px, and 128x128px size images}

The images were set to a standard of 512x512px size, which was the maximum size possible with the available dataset. The final dataset consisted of 10070 representative images of thin sections in both XPL and PPL from four different classes, (i) plutonic, (ii) volcanic, (iii) metamorphic, and (iv) sedimentary rocks. The initial model was evaluated using the FID score when it reached 80 Kimgs to assess the training speed.  The following model was evaluated every 140 Kimgs processed. Additional models were trained using 256x256px and 128x128px versions of the dataset, with the purpose of evaluating how well did the FID Score performed under different resolutions, while keeping the same dataset size. Training was terminated when the values did not improve and started oscillating, i.e., convergence, the model with the lowest FID Score achieved being selected. Training was conducted using a Quadro RTX 5000 with 16 Gb of video RAM, 30Gb of RAM and an eight-core GPU taking (1) 264 GPU hours  for the 512x512px model, (2) three days and five hours for the 256x256px model and (3) 72 GPU hours for the 128x128px model. 

\subsubsection*{Training procedure using style-mixing on top of the 512x512px model}

To evaluate the capability of the model to adapt to specific lithologies. We tested the 512x512 model as a starting step for generating domain-specific thin section models, with this in mind we used three lithology classes from the original dataset, using data augmentation and slicing on the original dataset available. The main aim is to generate the highest number of domain specific lithologies without the limitation of class balancing, previously used in the all-lithology model. Training was resumed using the 512x512 all-lithology model and trained during 1120 Kimgs, and it lasted for 44 hours with the GPU used. 

\subsection*{Survey with domain subject experts and students}

\begin{figure}[ht]
\centering
\includegraphics[width=\linewidth]{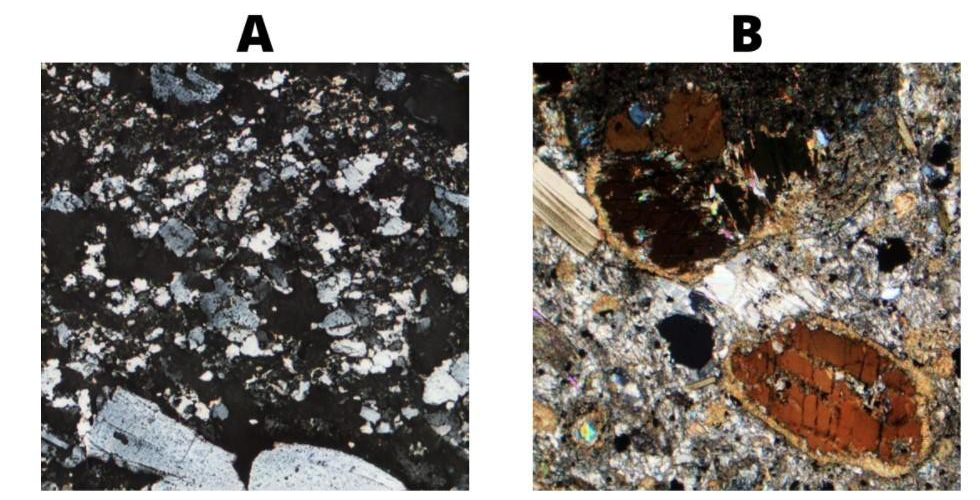}
\caption{First question presented on the survey. Two images side by side with three answers options, the right image (B) being the real one in this case.}
\label{fig:fig4}
\end{figure}

As an additional step for evaluating the performance of the final model, a survey was made to assess if the generated thin sections were indistinguishable from the real ones; this survey was aimed at subject matter experts from academia and industry with both geoscience and non-geoscience backgrounds, globally. In the survey, ten actual petrographic images were selected at random from the training dataset and ten randomly generated artificial images with randomly selected seed numbers were compared. In addition, the location of the correct image was also randomized. For example, the correct image is on the right and corresponds to an Aillikite from the Strekeisen dataset, and the generated one to the left corresponds to seed 0008 in the model as seen in Figure \ref{fig:fig4}. In total, more than two hundred responses were received during a short three-day survey.

\section*{Results}

\subsection*{Model Performance evaluation using FID Score}

\begin{figure}[ht]
\centering
\includegraphics[width=\linewidth]{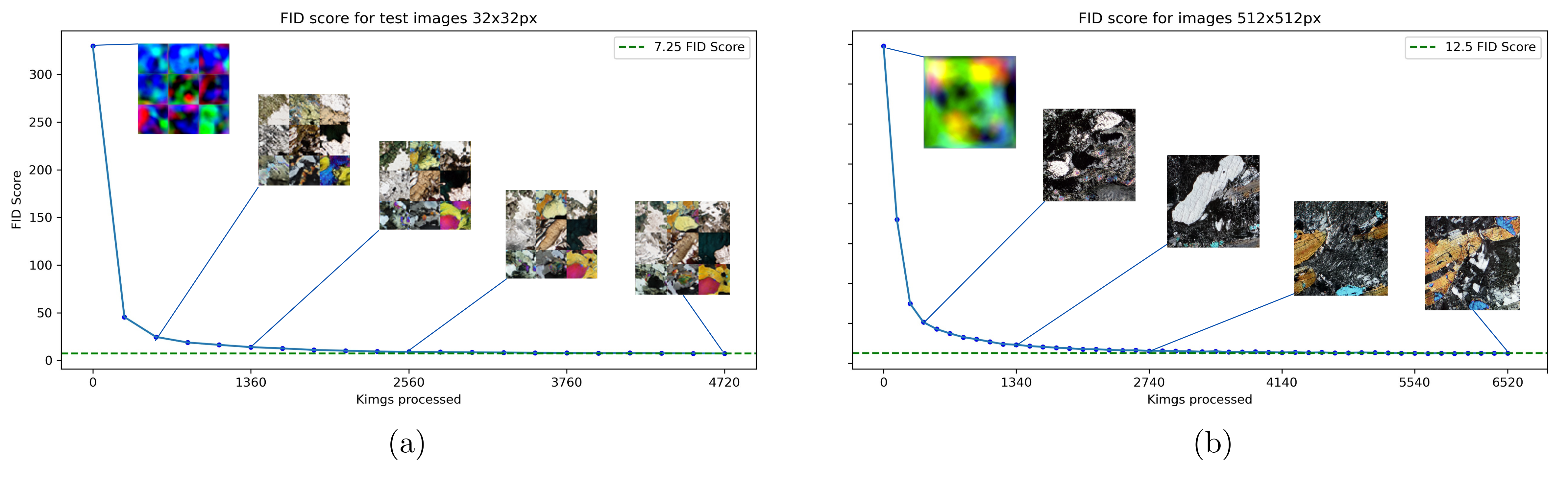}
\caption{Fréchet Inception Score evolution for (a) the 32x32 training, it converges around 7.24 FID score, a 3x3 grid of images evolution during training is shown and (b) evolution of the FID Score for the 512x512 main training, it converges around 12.5 FID Score}
\label{fig:fig5}
\end{figure}

The FID score obtained for the reduced size 32x32 px of the dataset was low compared with other resolutions and the final FID score obtained was 7.5 for this dataset, Figure \ref{fig:fig5}a. A timelapse of the generated images for the 32x32px and 512x512px models is shown in Figure \ref{fig:fig5}, showing the evolution of a 3x3 grid of images from noise to low-resolution artificial thin sections in the 32x32px model and a single thin section in the 512x512px model. For each 240 Kimgs processed, the FID score was evaluated for the 32x32px pixel dataset, and each 140 Kimgs it was evaluated for the 512x512px pixel image dataset. 
As stated in the methods section and after proving with the MVP that a generative model using StyleGAN2 was feasible, the network was trained with 512x512px resolution images, the FID score obtained was 12.49, Figure \ref{fig:fig5}b,. This is encouraging because as far as the literature review done for this paper, this score is the state-of-the-art FID score for a GAN model trained on microphotographs encompassing all three lithologies. In the training, the FID score stabilized at around 2740 Kimgs processed, and no significant increase was observed after 6520 Kimgs; hence, we obtained the lowest FID score achieved as the final model.

The final model was used to train specific petrographic groups of thin sections of different dataset sizes. Using it as a way of transfer learning, style mixing in the context of GANs, training was stopped at 1120 Kimgs for each lithology compared to the 6520 Kimgs reached by the original model. Different lithologies and image sizes were trained, a summary of training time provided in Table \ref{tab:table4}.

\begin{table}[htbp]
        \centering
        \setlength\tabcolsep{2pt}
\begin{tabular}{|c|c|c|c|c|c|}
\hline
Lithology & Resolution & Dataset size & Training Kimgs & FID Score & GPU Used \\
\hline
All lithologies & 512x512 & 10070 & 6520 & 12.49 & RTX5000\\
\hline
Sedimentary* & 512x512 & 5995 & 1120 & 24.19 & RTX5000\\
\hline
Metamorphic* & 512x512 & 13325 & 1120 & 14.40 & RTX5000\\
\hline
Igneous* & 512x512 & 11350 & 1120 & 16.11 & RTX5000\\
\hline
All lithologies & 256x256 & 10070 & 6160 & 11.89 & RTX5000\\
\hline
All lithologies & 128x128 & 10070 & 8320 & 10.41 & RTX5000\\
\hline
Igneous plutonic & 32x32 & 15294 & 4720 & 7.24 & M4000\\
\hline
\end{tabular}
\caption{\label{tab:table4}Value of the FID score obtained on different lithologies and dataset size. Models in boldface were trained from scratch *models trained using transfer learning applied to the all-lithologies 512x512 model}
\end{table}

\subsection*{Synthetic petrographic images}

\begin{figure}[ht]
\centering
\includegraphics[width=\linewidth]{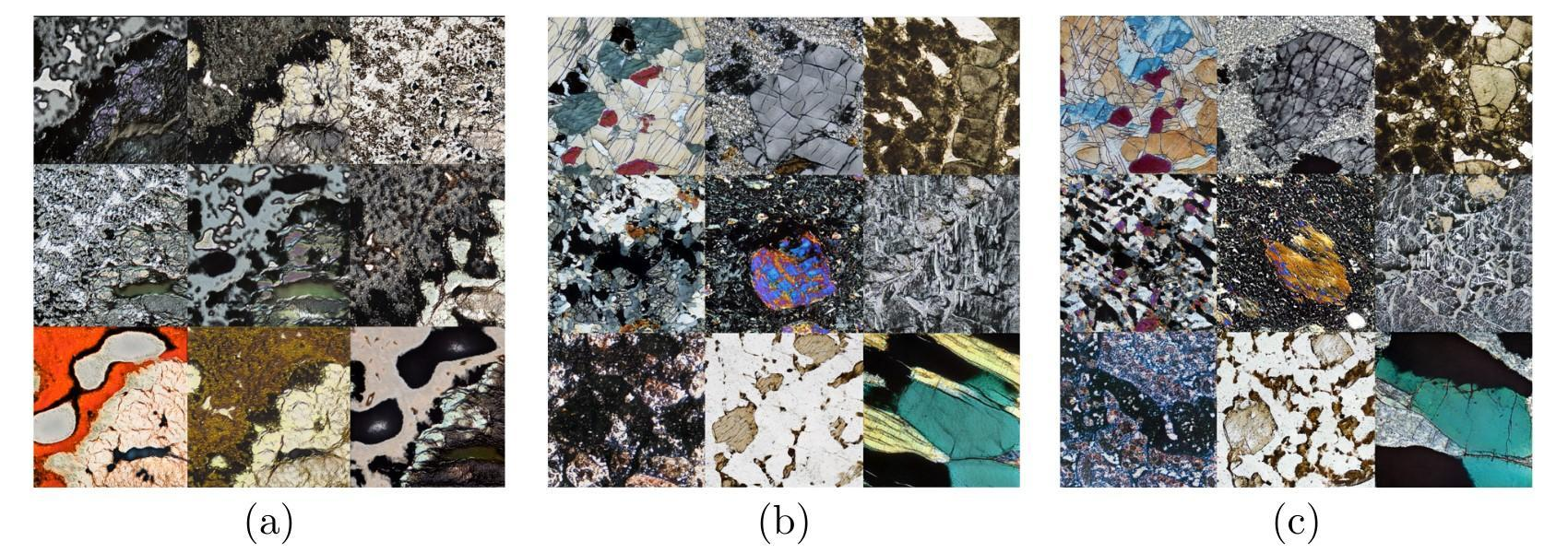}
\caption{Comparison of different stages of training for selected images same images generated at (a) 74.59 FID score, (b) 32.84 FID score, and (c) 12.49 FID score}
\label{fig:fig6}
\end{figure}

The images were generated in grids when the FID score was calculated, in the case of the 512x512 px model every 140Kimgs, evaluating the progressive improvement in the quality of the generated images, as shown in Fig 5. The GAN starts from random noise, and it progressively improves until it reaches convergence, i.e., the point where no further training would improve the model, as seen in Figure \ref{fig:fig5} The grid visualization also helps in spotting mode collapse whereby the generator becomes proefficient at producing one thin section and only generates variants of that image. Nine selected generated images are shown in Figure \ref{fig:fig6} with different FID scores during the training of the 512x512 model, the seeds were the same, and it shows a progressive improvement of mineral-like structures in the synthetic images.

\subsection*{Survey results}

\begin{figure}[ht]
\centering
\includegraphics[width=\linewidth]{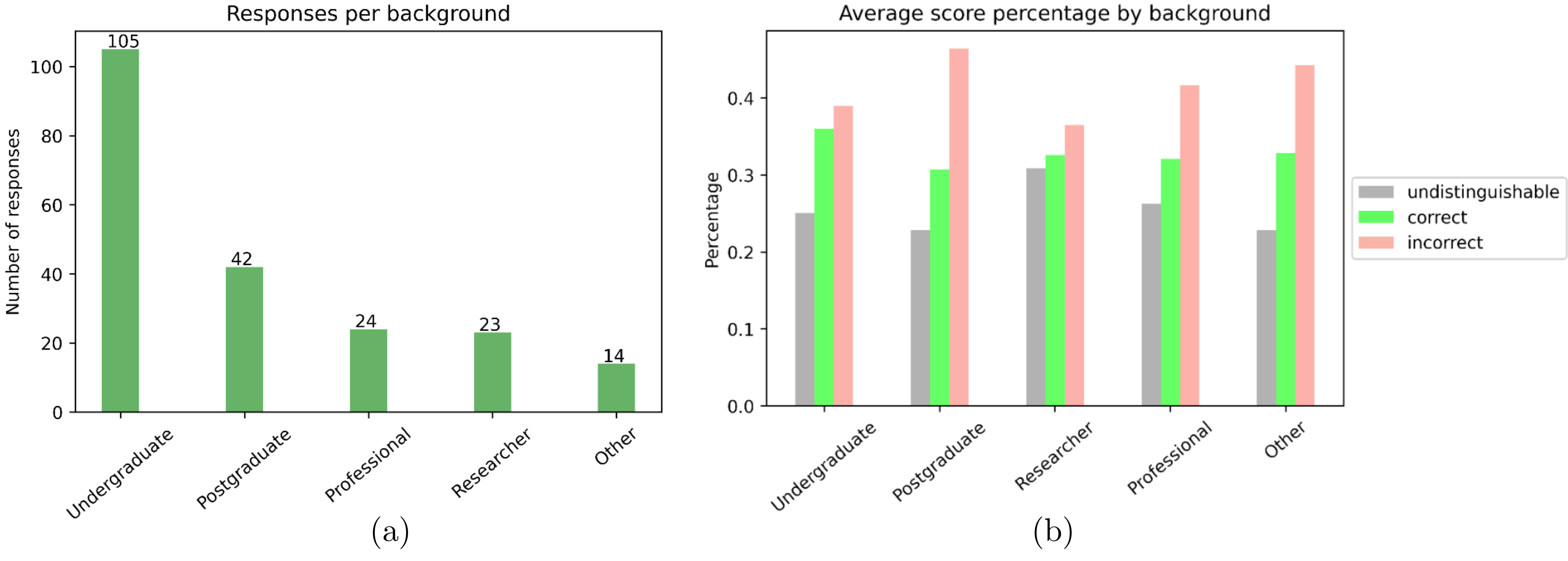}
\caption{(a) Academic or industry background for individuals with a geoscience related background and (b) results of the experiments, with average results by population background}
\label{fig:fig7}
\end{figure}

Results of the survey made with the purpose of evaluating the quality of the generated images are presented in Figure \ref{fig:fig7}. The survey was applied to 205 individuals worldwide from different backgrounds, in both industry and academia contexts, most of the responses come from undergraduate and postgraduate geoscience students, backgrounds shown in Figure \ref{fig:fig7}. 

Although the overall results of the survey in different backgrounds were similar, we observe that the performance of surveyees with different background,‘other’ in Figure \ref{fig:fig7}, is generally lower than those with a geoscience background. Across all background categories, undergraduate students have the highest performance, postgraduates have the lowest performance, and researchers have the highest percentage of doubts. Overall, the survey results show that on average the generated images perform better on all backgrounds.

\section*{Discussion}

\begin{figure}[ht]
\centering
\includegraphics[width=\linewidth]{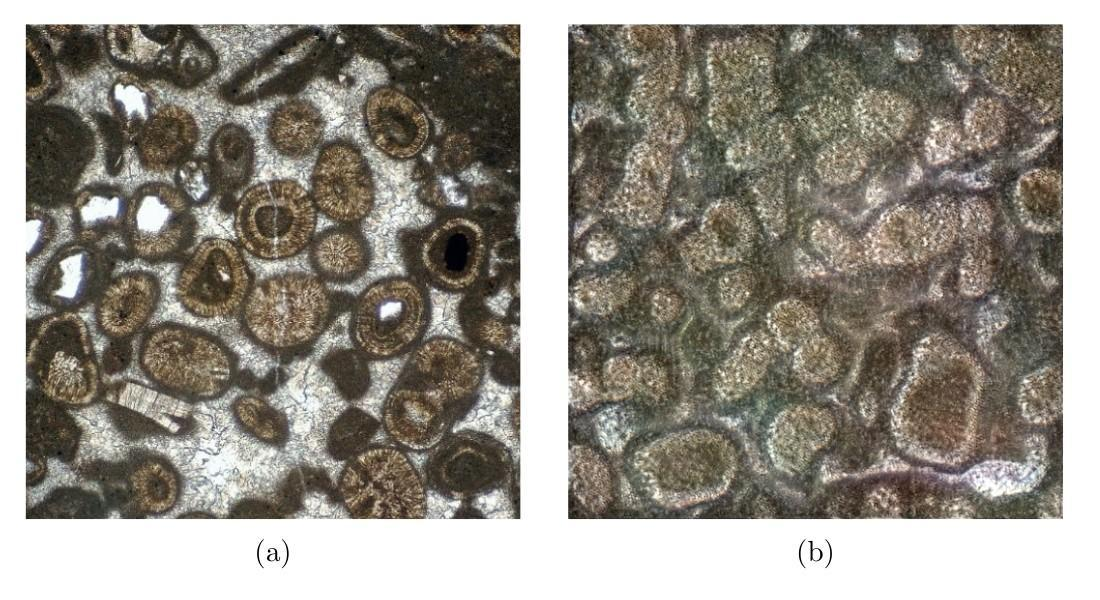}
\caption{Results of interpolating an image into the model (a) Ooilitc limestone \cite{Oxford2005} (b) interpolated oolitic limestone in the model’s latent space}
\label{fig:fig8}
\end{figure}

The proposed use of a GAN trained on geological data and in this case with petrographic images is the ability to visualize thin sections as a moving system, as this could be a way to picture the changing state of different lithologies, so far, an application of this is to give a real thin section not seen by the model during training and searching for the associated latent vector. This could lead to similar images found in the model.
An example of this shown as an ooilitic limestone taken from the University of Oxford Rocks Under the Microscope project \cite{Oxford2005} is given to the model, Figure \ref{fig:fig8}a, which then proceeds to search for the most similar image within its latent space, Figure \ref{fig:fig8}b. Resulting in a vector generated for the artificial image found within latent space, the proposed use of this feature is to search for similar thin sections and experiment with proximal vectors as a way of visually experimenting with the lithology.

Searching for a similar thin section in latent space could help us visualize how would a computer machine learning model organize a petrographic set of images, which sections would it tend to group together and which ones apart as a kind of taxonomy that it would use to group lithologies in latent space, and what features would be more dominant and how could we control the most important ones from a geological point of view, e.g. grain size or foliation, to generate specific textures.

\begin{figure}[ht]
\centering
\includegraphics[width=\linewidth]{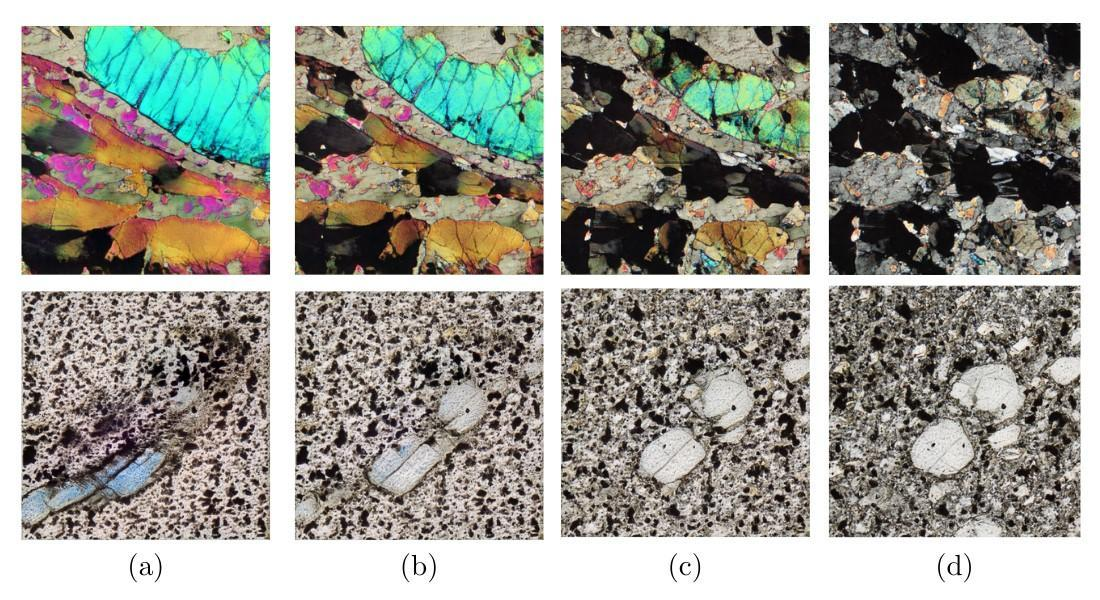}
\caption{Results of applying truncation to two generated images, with the intention of showing a progressive change in truncation and its effects, with (a) 1.25 (b) 1.0 (c) 0.75 (d) 0.5 as truncation values}
\label{fig:fig9}
\end{figure}

A modification of the latent distribution implementing a truncation of the normal distribution used to generate images, i.e., truncating the values which fall above a certain threshold; it is called the Truncation Trick \cite{Brock2018}. This has been shown to improve and boost the FID score of the generated images and was used in the survey as a way of increasing the probability of an artificial thin section to appear as a real one, using 0.7 as a truncation value. The experimentation with this truncation value is shown to produce more unrealistic minerals the greater the value of the threshold is used, generating images with varying threshold values shown in Figure \ref{fig:fig9}, while reducing the truncation value produces more down-to-earth minerals albeit with a tendency to make the general color of the thin section gray.

\begin{figure}[ht]
\centering
\includegraphics[width=\linewidth]{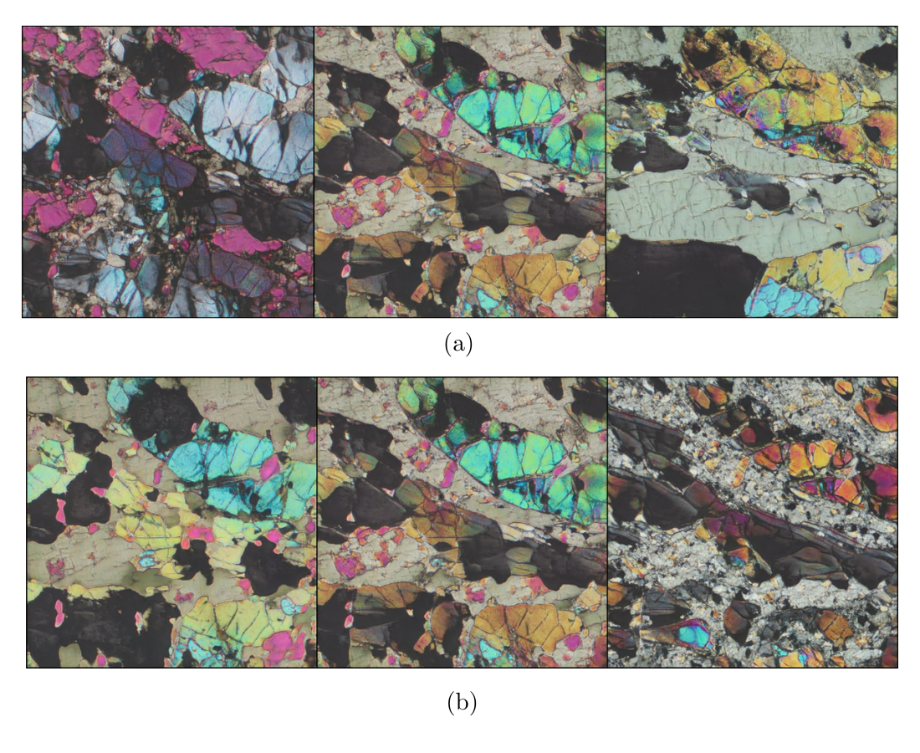}
\caption{Feature Extraction of important vectors in the model, the same seed, 168947, in the center is changed (a) modifying colors of the minerals and (b) changing the percentage of matrix in the thin section}
\label{fig:fig10}
\end{figure}

An application of being able to generate synthetic data is the ability to extract human-readable feature vectors in latent space, we used the Closed-Form Factorization \cite{Shen2020} of latent vectors for the all-lithologies 512x512 model, this method could be used in the future for visualizing different features being modified on the same mineral assemblage, Figure \ref{fig:fig10}. This showed that we could use the trained model to extract vectors that can be used to modify the same thin section and add or remove certain constituents. Future applications of this factorization could be petrographic and petrological modeling, especially if this vector can be associated with certain characteristics of geological environments, an interest application being grain size modification and kind of minerals present; this model could also be used to visualize facies and lithological changes and could assist in geological workflows that rely heavily on petrographic information.

We also apply this method to an image classification problem, using 200 images of landscapes and 200 artificial thin sections, we trained a deep convolutional neural network architecture and tested the model using images of landscapes and real thin sections, the model was able to reach 95\% accuracy with the training data and 80\% accuracy with the testing set.

Synthetic images were found to be more prone to be classified as real, Figure \ref{fig:fig7}, than real thin sections. This phenomenon could be explained because generated images tend to look more like an average thin section given that they are trained to assimilate an entire distribution of images; this “archetypical” thin section then would be erroneously classified as the real one when compared with a single real thin section in a binary classification task, i.e., real, or fake, when a human is used as the classifier. Images that are ‘more real than real” have already been observed in GANs trained with faces\cite{Lago2022,Nightingale2022} and Gestalt theory has been previously used in deep learning in preprocessing steps to obtain efficient image descriptors for CNN \cite{Horhan2020} training. With this study, we propose that this ‘gestalt’, i.e., the laws pertaining our ability to make meaningful perceptions of the world \cite{Shi2016}, GAN phenomenon could extend to geological nonfacial datasets and is something that should be considered and further studied, as this phenomenon could indicate some continuity, memory, similarity, closure and good figure in the sense of Gestalt theory regarding our understanding and perception of synthetic and real petrographic data; we made an attempt to address one of these Gestalt principles with a symmetry test between the real and fake images used for the survey, were found to have higher symmetry.

The significance of this model is enabling the generation of artificial thin sections and with further studies could be used as a viable method for dataset augmentation, with the potential as a tool for self-labeling being an input to semisupervised and unsupervised learning algorithms; explainability of this kind of model is also an area of research and could elucidate in the future how a GAN organizes data in its latent space. It is also noted and encouraged that the final model can be used as the starting point for training more domain-specific petrographic datasets, and this could be done through style-mixing of the GAN model, with the aim of generating more specific generative models, e.g., in the generation of carbonate constituents \cite{Koeshidayatullah2020} or an only metamorphic thin section generator. 

This architecture also makes it possible for the images to be generated according to a signal, an implementation of this being the audio reactive GAN "MAUA" implementation \cite{Brouwer2020}.Further exploration and evaluation of the generated thin sections in latent space could aid in the evaluation of how a given lithological feature evolves and in the future, this could be used to assist in interactive explanation and visualization or in modeling of petrographic environments e.g., the impact of varying levels of metamorphism on a thin section and the effects of change in energy levels in a sedimentary environment. We also observe that, with the different image sizes tested we expect to get lower scores, i.e. better, for smaller image sizes, with a comparison given in Table 4, we can get an idea of the dataset size needed to achieve a target FID Score. 

\section*{Future Recommendations}

We encourage the implementation of the recently released (at the time of writing) StyleGAN3 model and upcoming architectures to further improve the current model and the use of the trained model in more domain-specific datasets. Exploration of latent space and feature modification of thin sections is needed as ways to prove that this type of architecture will help in the visualization of changing variables in geological environments by the way of changes in latent space, Image-to-Image translation is suggested to generate petrographic images from another type of images, and an implementation of super-resolution \cite{Niu2020} would be most needed to up-sample available petrographic datasets resolutions.

Exploration of features extracted from the model is a way forward to control certain geological characteristics of the generated data, i.e., a feature for controlling the grain size, the predominance of the matrix over grains, or the abundance of a certain mineral species. It is also recommended to explore ways to associate latent vectors with geochemical data, to be able to visualize the effects of changing modal composition on a thin section; this could be useful for example, to generate thin sections based on modal composition in metamorphic petrology modeling. A more discrete survey is advised, i.e., generating a model trained on a specific lithology, thus enabling more domain-specific tests to be made, e.g., assessing sedimentologists or petrologists to give an artificial thin section tentative metamorphic or sedimentary facies. 
We tested the GAN model capacity as a tool to generate datasets for other machine learning algorithms, for this we trained a binary image classifier using a Convolutional Neural Network over 100 synthetic thin sections versus 100 landscape images, the model achieved over 90\% accuracy on training and testing and when tested against 40 real thin sections, the accuracy dropped, but was over 80\% nonetheless, further validation is needed to use this kind of model as a data augmentation tool in future geoscience workflows.

\section*{Conclusions}

\begin{enumerate}
  \item It is possible to generate an artificial dataset of petrographic thin sections using Generative adversarial networks, via the architecture of StyleGAN2. The training of a viable GAN using StyleGAN2 in this context needs at least 5000 images to achieve sufficiently good images, and more than 10000 images is recommended to generate an optimal model (i.e., lower than 15 FID score). 
  \item Based on the result of the survey, we conclude that artificially generated thin sections can be indistinguishable from real ones and even be seen as more authentic than real ones, allowing this tool to generate thin sections of sufficient quality to be able to deceive domain subject experts.
  \item Latent space exploration of the model is a method of visualization and interpolation of real thin sections into the model. Further exploration of styles in the context of petrography is needed to support GAN models as a technique for petrographic modeling.
  \item Closed form factorization of latent space in a petrographic image generator is used for extracting at least two human-readable vectors that could be used in the future for modeling purposes in the geosciences. 
\end{enumerate}

\bibliography{main}

\section*{Acknowledgements}

We thank Dr. Alessandro Da Mommio from the University degli Studi di Milano and Dr. Nathan Daczko from the University of Macquarie for the collected images made publicly available at their respective sites. Thanks to the people at Paperspace, especially Jordan Burke for the help provided during training troubleshooting and James Skelton for the wonderful exposition of the cloud service features. Thanks to all students, professors, and professionals who helped with the survey; answers to the real thin sections made in the test are attached as an annex. Thanks to the members of the Society of Economic Geologists student chapter from the National University of Colombia in Bogotá and the King Fahd University of Petroleum and Minerals for the divulgation efforts and support of this work.

\section*{Author contributions statement}

I.F. model training, creation of tests, and manuscript writing  A.K. original idea, manuscript writing.  All authors reviewed the manuscript. 

\section*{Additional information}

\subsection*{Availability of Data and Materials}

 The dataset and code used and/or analysed during the current study available from the corresponding author on reasonable request. This is a manuscript under review process and the trained models will be available soon. For the StyleGAN2+ADA implementation please refer to \href{https://github.com/NVlabs/stylegan2-ada-pytorch}{https://github.com/NVlabs/stylegan2-ada-pytorch}

\section*{Competing interests}

The authors declare no competing interests.

\end{document}